\documentclass[letterpaper, 10 pt, conference]{ieeeconf}  

\IEEEoverridecommandlockouts                              

\usepackage{graphicx}
\usepackage{makecell}

\title{\LARGE \bf
Improving Omics-Based Classification: The Role of Feature Selection and Synthetic Data Generation
\thanks{© 20XX IEEE.  Personal use of this material is permitted.  Permission from IEEE must be obtained for all other uses, in any current or future media, including reprinting/republishing this material for advertising or promotional purposes, creating new collective works, for resale or redistribution to servers or lists, or reuse of any copyrighted component of this work in other works.}
}

\author{Diego Perazzolo$^{1,2}$, Pietro Fanton$^{2}$, Ilaria Barison$^{1}$, Marny Fedrigo$^{1}$, Annalisa Angelini$^{1}$, \\ Chiara Castellani$^{1**}$ and Enrico Grisan$^{3*}$
\thanks{$^{1}$Department of Cardiac, Thoracic, Vascular Sciences and Public Health, University of Padova, Via Giustiniani, 2 - 35128 Padova, Italy
        {\tt\small diego.perazzolo.1@phd.unipd.it}, {\tt\small ilaria.barison.1@studenti.unipd.it}, {\tt\small marny.fedrigo@aopd.veneto.it}, {\tt\small annalisa.angelini@unipd.it}, {\tt\small chiara.castellani@unipd.it}}%
\thanks{$^{2}$I4 Consulting Srl,
        Galleria Milano, 1 - 35139 Padova, Italy
        {\tt\small pietro.fanton@i4consulting.it}}%
\thanks{$^{3}$School of Engineering, London South Bank University,
        103 Borough Road, London, SE1 0AA, UK
        {\tt\small enrico.grisan@lsbu.ac.uk}}%
\thanks{*Corresponding Author}%
\thanks{**Co-Corresponding Author}
}

\begin{document}

\maketitle
\thispagestyle{empty}
\pagestyle{empty}

\begin{abstract}
Given the increasing complexity of omics datasets, a key challenge is not only improving classification performance but also enhancing the transparency and reliability of model decisions. Effective model performance and feature selection are fundamental for explainability and reliability. In many cases, high-dimensional omics datasets suffer from limited number of samples due to clinical constraints, patient conditions, phenotypes rarity and others conditions. Current omics-based classification models often suffer from narrow interpretability, making it difficult to discern meaningful insights where trust and reproducibility are critical. This study presents a machine learning-based classification framework that integrates feature selection with data augmentation techniques to achieve high-standard classification accuracy while ensuring better interpretability. Using the publicly available dataset E-MTAB-8026, we explore a bootstrap analysis in six binary classification scenarios to evaluate the proposed model’s behaviour. We show that the proposed pipeline yields cross-validated perfomance on small dataset that is conserved when the trained classifier is applied to a larger test set. Our findings emphasize the fundamental balance between accuracy and feature selection, highlighting the positive effect of introducing synthetic data for better generalization, even in scenarios with very limited samples availability. 
\newline

\indent \textit{Clinical relevance}— The proposed framework addresses a critical challenge in omics-based disease classification by improving model interpretability while maintaining high classification performance. By providing a more explainable data-driven approach, this study contributes to the development of reliable and reproducible diagnostic tools that can support clinical decision-making.
\end{abstract}

\section{INTRODUCTION} 
Advancements in omics technologies have facilitated large-scale profiling of molecular biomarkers, offering unprecedented insights into disease mechanisms and therapeutic targets. However, despite the vast amount of data generated through high-throughput sequencing and microarray technologies, the effective utilization of omics data remains a significant challenge. One of the primary obstacles arises from the high dimensionality of these datasets, often coupled with limited sample sizes, leading to an inherent imbalance that complicates classification tasks and increases the risk of overfitting. Furthermore, this imbalance can influence the interpretability of model decisions, making it difficult to derive meaningful insights from the classifier's outputs \cite{lay2006problems} \cite{kang2022roadmap}.
Machine learning (ML) has emerged as a powerful tool in omics research, enabling the identification of meaningful patterns and enhancing diagnostic accuracy \cite{sidak2022interpretable}. ML has been used to develop diagnostic, prognostic and predictive tools from single omics data and now also being applied to multi-omics data to investigate and interpret the relationships \cite{reel2021using}. Feature selection techniques play a crucial role, especially in biological data analysis, in reducing model complexity by identifying the most relevant features, thereby improving generalization to new datasets and ensuring more interpretable decision-making \cite{maitra2023umint}. Data augmentation techniques, offer a promising approach to address the limitation of low-samples availability. By generating synthetic data that mimics real patterns, augmentation can enhance model robustness and improve performance in scenarios with small sample sizes \cite{ahmed2022multi}.
In this study, we introduce a machine learning-based classification framework, that combines LASSO-based feature selection \cite{tibshirani1996regression}, data augmentation techniques and KSVM model \cite{scholkopf2002learning}, to enhance both model performance and interpretability. Leveraging the publicly available dataset E-MTAB-8026 on ArrayExpress functional genomic data collection, we conduct a bootstrap analysis across six binary classification scenarios, with the aim to assess the impact of data availability during training, introduction of synthetic data generation procedures on omics-based classification, accuracy and feature selection of the proposed model. Our findings underscore the critical tradeoff between feature selection and accuracy, revealing that while a baseline LASSO-based models retaining all nonzero coefficients achieve slightly higher accuracy with respect to our proposed framework, features selection and incorporation data augmentation leads to a more interpretable model wit comparable accuracy. By balancing these factors, our approach aims to support the development of robust, explainable decision support systems, facilitating omics data exploration and biomarker discovery for personalized treatments.

\begin{figure*}[thpb]
    \centering
    \includegraphics[width=0.85\textwidth]{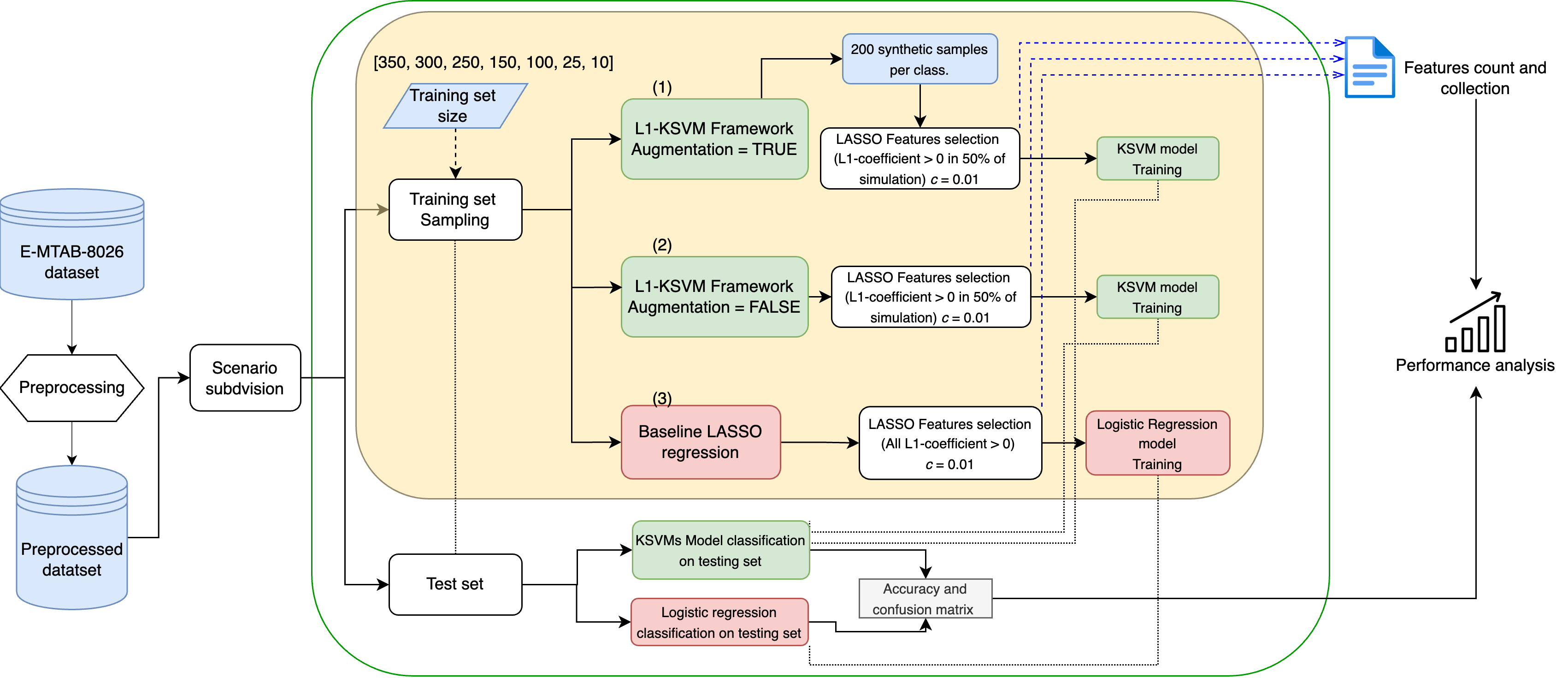}
    \caption{Experimental Procedure for the Bootstrap Analysis. The preprocessed dataset is divided into six binary classification scenarios. For each scenario, training set sampling is performed at different sizes, while the remaining samples constitute the test set. Three classification approaches are evaluated: (1) the L1-KSVM Framework with augmentation, where 200 synthetic samples per class are generated using a Gaussian noise-based augmentation technique, (2) the L1-KSVM Framework without augmentation, which follows the same feature selection and classification procedure without synthetic data, and (3) the Baseline LASSO regression model, where features are retained if their L1-coefficient is greater than zero. Features collection and classification is then performed. Resulting data are collected to conduct performance analysis.}
    \label{bootstrap_analysis_schema}
\end{figure*}

\section{MATERIALS AND METHODS}

\subsection{Dataset Description}
The dataset used in this study was obtained from the publicly available ArrayExpress repository (E-MTAB-8026), which contains circulating microRNAs (miRNAs) expression profiles associated with lung cancer detection in symptomatic patients. The study and all experimental procedures involving human subjects were approved by the Ethics Committee of the \"{A}rztekammer des Saarlandes (\"{A}rztekammer Saarland, Saarbr\"{u}cken, Germany), and verbal informed consent was obtained from all participants \cite{fehlmann2020evaluating}. miRNA expression profiles derived from whole blood samples of patients who underwent clinical evaluation for lung cancer. Blood samples were collected from patients with confirmed lung cancer (LCa), non-tumor lung diseases (NTLD), other diseases (OD), and healthy controls. Total RNA was extracted using Qiagen PAXgene Blood miRNA kit, labeled and hybridized using Agilent miRNAComplete Labeling and Hyb kit. Agilent microarray scanner system have been used to generate miRNAs expression profiles. The dataset includes a total of 3,140 samples, divided in the following classes:
\newline
\begin{itemize}
    \item \textbf{Lung Cancer (LCa):} 606 samples
    \item \textbf{Non-Tumor Lung Disease (NTLD):} 593 samples
    \item \textbf{Other Diseases (OD):} 977 samples
    \item \textbf{Healthy Controls:} 964 samples
\end{itemize}

\subsection{Data Preprocessing}
miRNAs with invalid expression values were removed to reduce noise and improve data quality. Only microRNAs with \textit{hsa-} prefix were retained, ensuring that the dataset exclusively contained human-associated miRNAs. After this filtering steps, The resulting dataset included the expression of 1,184 miRNAs as features. 
To create a well-balanced dataset while reducing computational complexity, we selected an equal number of 500 samples per class (LCa, NTLD, OD, and Healthy Control). This balanced subset ensures that the classification model is not biased toward any particular group while maintaining sufficient statistical power for analysis.

\subsection{Model Architecture}
The proposed model classification framework design, consists in the application of 3 main steps. 
\newline
\subsubsection{Synthetic Samples Generation}
To mitigate the cases of data scarcity and enhance model robustness, a Gaussian noise-based augmentation strategy is applied \cite{bishop1995training} to the training data to generate synthetic samples while preserving the original data distribution. The process involves selecting real samples from each class and perturbing their values with Gaussian noise, where the standard deviation of the noise is set as the 10\% of the original feature’s standard deviation. For each feature, we independently compute the standard deviation within each class, ensuring that variations reflect intrinsic variability. New synthetic samples are generated by randomly selecting values from the original dataset  and adding the computed noise. This method maintains the statistical properties of the dataset while introducing controlled variability to improve model generalization. The final augmented dataset includes an equal number of synthetic samples per class. 
\newline
\subsubsection{Feature Selection}
To tackle high-dimensional datasets, a feature selection approach based on L1-regularized logistic regression (LASSO) is applied. This method penalizes feature coefficients, effectively reducing the weights of irrelevant features to zero, thereby selecting only the most influential ones \cite{muthukrishnan2016lasso}. Multiple LASSO simulations are conducted, each trained on a dataset augmented with synthetic samples. Inverse regularization strength parameter ($c$) have been setted to 0.01. After each simulation, features with non-zero L1 coefficients are recorded. 
The occurrence of each selected feature across multiple simulations is counted, and only those present in more than 50\% of the executed simulations are retained and used as input features for the subsequent classification step.
\newline
\subsubsection{Kernel-Based Classifier}
Kernel Support Vector Machine (KSVM) model with polynomial kernel is employed. KSVM is a powerful method that exploit a kernel function to project data into a higher-dimensional space, defining the optimal hyperplane that maximally separates the two classes \cite{scholkopf2002learning}. The formulation can be denoted from the equation (\ref{KSVM_equat}), where \( \alpha_i \) are the learned weights, \( y_i \) are the class labels, and \( x_i \) are the support vectors. The function \( K(x_i, x) \) is the kernel function that maps data into a higher-dimensional space, and \( b \) is the bias term that shifts the decision boundary. $n$ is the number of support vectors and $f(x)$ is the decision function that determines the classification of $x$. The classifier is trained using the selected features from the feature selection step, ensuring that only the most relevant ones contribute to the decision boundary. Once trained, the KSVM predicts class labels for the original test samples, and performance is evaluated using standard metrics such as accuracy and confusion matrices.

\begin{equation}
	f(x) = \sum_{i=1}^{n} \alpha_i y_i K(x_i, x) + b
	\label{KSVM_equat}
\end{equation}
For simplicity, we refer to the proposed architecture as the L1-KSVM Framework throughout this study.

\subsection{Experimental procedure}
The experimental procedure is designed to evaluate the effect of sample size on classification performance and the interpretability of feature selection. A visual representation of the process is provided in Figure \ref{bootstrap_analysis_schema}. We conducted a bootstrap analysis on six binary classification scenarios. 
\newline
\begin{itemize}
    \item \textbf{Scenario 1:} Healthy Controls vs LCa
    \item \textbf{Scenario 2:} Healthy Controls vs NTLD
    \item \textbf{Scenario 3:} Healthy Controls vs OD
    \item \textbf{Scenario 4:} LCa vs NTLD
    \item \textbf{Scenario 5:} LCa vs OD
    \item \textbf{Scenario 6:} NTLD vs OD\\
\end{itemize}
For each predefined classification scenario, we assessed different training set sizes by randomly selecting subsets from the balanced dataset, which consists of 500 samples per class. The training set sizes tested were: 350 (70\%), 300 (60\%), 250 (50\%), 150 (30\%), 100 (20\%), 25 (5\%), and 10 (2\%). All remaining samples constituted the test set. Each experiment was repeated 100 times per sample size, with new bootstrap samples drawn in each iteration. Three classification approaches were evaluated:
\begin{itemize}
    \item L1-KSVM Framework incorporating 200 synthetic samples per class, generated using a Gaussian noise-based augmentation technique.\\
    \item L1-KSVM Framework without data augmentation, applying the same feature selection and classification procedure but without synthetic data.\\
    \item A baseline LASSO logistic regression model, where features were retained solely based on having a nonzero L1 coefficient.\\
\end{itemize}
The trained models were tested on the test set, and classification performance is measured using accuracy and confusion matrices. Counts of the microRNAs (features) selected by the regularization method are collected for each sample size across the 6 classification scenarios. The final results are analyzed to determine the effectiveness of proposed methodology. 
To avoid overestimating the performance of the proposed framework, we also conducted a cross-validation analysis for each sample size and computed the average accuracy across the six classification scenarios. This approach emulates a real-world scenario and permits determining if the accuracy is conserved when the trained classifier is applied to a larger dataset.


\begin{figure*}[th]
    \centering
    \includegraphics[width=0.95\textwidth]{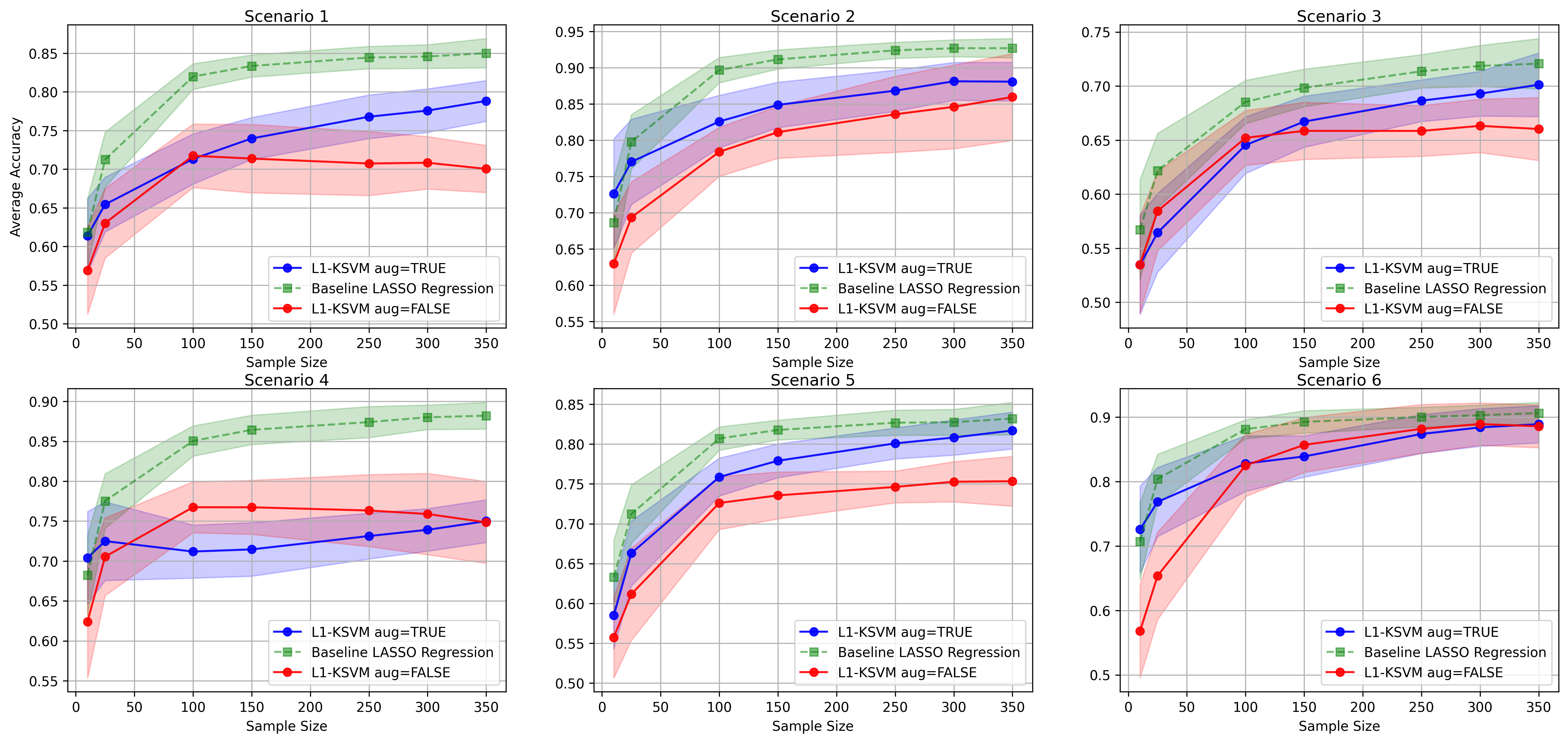}
    \caption{Each subplot corresponds to a binary classification scenario: Scenario 1 (Healthy Controls vs. LCa), Scenario 2 (Healthy Controls vs. NTLD), Scenario 3 (Healthy Controls vs. OD), Scenario 4 (LCa vs. NTLD), Scenario 5 (LCa vs. OD), and Scenario 6 (NTLD vs. OD). The x-axis represents the training sample size, while the y-axis shows the classification accuracy. The plots compare three approaches: the L1-KSVM Framework with Data Augmentation (blue solid line), incorporating 200 synthetic samples per class; the L1-KSVM Framework without Augmentation (red solid line), applying the same feature selection and classification process but without synthetic data; and the Baseline LASSO Regression Model (green dashed line), which retains all features with a nonzero L1 coefficient. Shaded areas indicate standard deviation across 100 bootstrap iterations per sample size.}
    \label{accuracy_analysis}
\end{figure*}

\section{RESULTS}
Figure \ref{accuracy_analysis} illustrates the classification accuracy across six binary classification scenarios for different training set sizes and classification methods. As expected, an increase in training sample size generally improves classification performance. Table \ref{performance_metrics} provides a detailed breakdown of performance metrics, including accuracy (Acc. \%), true positives (TP \%), true negatives (TN \%), false positives (FP\%), false negatives (FN \%) and the accuracy collected from the cross validation analysis (Cross-Val Acc. \%)

\begin{table}[h]
\caption{average number of selected micrornas across different classification scenarios for each training set size.}
\label{features_selection}
\begin{center}
\begin{tabular}{|c||c|c|c|}
\hline
\makecell{Training \\ set size} & \makecell{L1-KSVM \\ Aug. \\ (\# miRNAs)} & \makecell{L1-KSVM \\ without Aug. \\ (\# miRNAs)} & \makecell{Baseline \\\ LASSO reg. \\ (\# miRNAs)} \\
\hline
10&66.54 &5.68 & 1180\\
\hline
25&79.91 &10.33 & 1183\\
\hline
100&96.41 &19.93 & 1183\\
\hline
150&99.56 &20.39 & 1183\\
\hline
250&103.23 &20.72 &1183\\
\hline
300&104.56 &20.72 & 1183\\
\hline
350&107.05 &20.75 & 1183\\
\hline
\end{tabular}
\end{center}
\end{table}

The Baseline LASSO regression model (Baseline LASSO reg.) achieves relatively higher accuracy in multiple cases, resulting in an overall better performance across the six scenarios. However, Table \ref{features_selection}, which summarizes the average number of selected microRNAs (\# miRNAs) per training sample size, highlights a key advantage of the L1-KSVM-based approaches, their ability to perform feature selection efficiently. Compared to the Baseline LASSO regression, which retains a substantially larger number of features, both L1-KSVM approaches significantly reduce feature dimensionality, making them more interpretable and easier to investigate.

\begin{table}[h]
\centering
\caption{performance metrics across sample sizes for each method}
\label{performance_metrics}
\setlength{\tabcolsep}{5pt} 
\renewcommand{\arraystretch}{0.92} 
\begin{tabular}{|c|c|c|c|c|c|c|}
\hline
\textbf{\makecell{Sample \\ Size}} & \textbf{\makecell{Acc. \\ \%}} & \textbf{TP\%} & \textbf{TN\%} & \textbf{FP\%} & \textbf{FN\%} & \textbf{\makecell{Cross-Val \\ Acc. \%}} \\ 
\hline
\multicolumn{7}{|c|}{\textbf{Baseline LASSO Regression}} \\
\hline
10  & 64.9  & 32.74  & 32.16  & 17.26  & 17.84 & 61.2\\ 
25  & 73.7  & 37.08  & 36.64  & 12.92  & 13.35 & 71.7\\ 
100 & 82.3  & 41.82  & 40.52  & 8.17   & 9.48  & 81.8\\ 
150 & 83.6  & 42.60  & 41.02  & 7.39   & 8.97 & 83.2\\ 
250 & 84.7  & 43.24  & 41.47  & 6.75   & 8.52 & 84.6\\ 
300 & 85.0  & 43.42  & 41.60  & 6.57   & 8.39 & 83.8\\ 
350 & 85.2  & 43.54  & 41.75  & 6.45   & 8.24 & 84.1\\ 
\hline
\multicolumn{7}{|c|}{\textbf{L1-KSVM (Aug=FALSE)}} \\
\hline
10  & 58.0  & 29.46  & 28.58  & 20.54  & 21.42 & 57.1\\ 
25  & 64.6  & 32.40  & 32.24  & 17.60  & 17.75 & 62.0\\ 
100 & 74.5  & 37.11  & 37.41  & 12.88  & 12.58 & 73.6\\ 
150 & 75.7  & 37.73  & 37.98  & 12.26  & 12.01& 76.3\\ 
250 & 76.5  & 38.26  & 38.28  & 11.74  & 11.71 & 76.7\\ 
300 & 76.9  & 38.25  & 38.72  & 11.74  & 11.27 & 79.1\\ 
350 & 76.7  & 38.14  & 38.64  & 11.85  & 11.35 & 80.0\\ 
\hline
\multicolumn{7}{|c|}{\textbf{L1-KSVM (Aug=TRUE)}} \\
\hline
10  & 64.8  & 33.34  & 31.48  & 16.65  & 18.52 & 66.1\\ 
25  & 69.0  & 35.03  & 34.05  & 14.96  & 15.94 & 64.5\\ 
100 & 74.7  & 36.76  & 37.95  & 13.23  & 12.04 & 70.4\\ 
150 & 76.4  & 37.62  & 38.84  & 12.37  & 11.15 & 72.5\\ 
250 & 78.8  & 38.86  & 39.94  & 11.13  & 10.05 &74.5\\ 
300 & 79.6  & 39.40  & 40.28  & 10.59  & 9.71 & 77.2\\ 
350 & 80.4  & 39.82  & 40.61  & 10.17  & 9.39 & 79.2\\ 
\hline
\end{tabular}
\end{table}

Notably, the L1-KSVM framework with data augmentation improves classification accuracy over its non-augmented counterpart, achieving performance comparable to the Baseline LASSO regression, particularly in scenarios with limited training data. It is also possible to notice the positive effect of data augmentation, particularly evident in scenarios with limited training data, where it enhances model robustness and helps mitigate performance degradation. It is also possible to notice that the accuracy performances of the proposed framework,  obtained from the cross-validation analysis of each training sample size, are aligned with the accuracy of the same trained model applied to a larger dataset.

\section{DISCUSSIONS AND CONCLUSIONS}
The results of this study demonstrate the effectiveness of integrating feature selection and synthetic data generation in omics-based classification tasks. The baseline LASSO regression model exhibited relatively better classification accuracy across multiple scenarios, as shown in Figure \ref{accuracy_analysis} and Table \ref{performance_metrics}. However, this came at the critical cost of retaining a significantly larger number of features (Table \ref{features_selection}), which can reduce model interpretability and increase the risk of overfitting, particularly in high-dimensional datasets. The proposed L1-KSVM framework offered an optimal balanced trade-off between high accuracy and feature selection. The use of data augmentation significantly enhances classification performance compared to its non-augmented counterpart, particularly in low sample conditions, where the model benefits from the additional synthetic training examples, improving its generalization ability. Moreover, the L1-KSVM framework achieved comparable accuracy to the baseline model while drastically reducing the number of selected features, making it a more interpretable and explainable solution. 
We decided to include in the framework the L1-KSVM architecture to evaluate a more expressive model capable of capturing non-linear patterns through kernel-based projections while maintaining feature selection through L1 regularization. This combination aimed to balance model complexity and interpretability. The use of synthetic data via Gaussian noise further tested whether data augmentation could improve robustness in low-sample settings.
We demonstrate that the proposed frameworks achieve consistent cross-validated performance on small datasets that emulate real-world scenarios. The accuracy remains stable when the trained classifier is applied to a larger test set, highlighting its robustness and generalizability. Reducing high dimensionality by selecting relevant features, which enables accurate classification of patient conditions, provides a significant advantage in understanding underlying biological insights and facilitates more precise statistical and bioinformatics analyses. This study advances the integration of synthetic data into omics-based analysis while improving decision interpretability through targeted feature selection. Future research will aim to validate the framework across multiple omics datasets and explore more sophisticated data augmentation techniques, such as generative deep learning models (like generative adversarial networks (GANs) and variational autoencoders (VAE)) to design more biologically meaningful synthetic samples. This approach will be crucial in overcoming the study's limitation, as Gaussian noise-based augmentation may not fully capture the complex biological variability present in real-world omics datasets.

\addtolength{\textheight}{-12cm}   






\bibliographystyle{IEEEtran}  
\bibliography{IEEE_classifier_biblio}  

\end{document}